\documentstyle[12pt,a4,epsf]{article}

\makeatletter
\def\@cite#1#2{{#1\if@tempswa , #2\fi}}
\def\@biblabel#1{\hspace*{5mm}\hfill}
\makeatother

\begin{document}
\begin{titlepage}
\title{
\vspace*{2cm}
{\Large \bf Word Length Frequency and Distribution in
    English:\\ \vspace*{5mm}
    \large \bf Observations, Theory, and Implications for the
    Construction of Verse Lines}}
\author{Hideaki {\sc Aoyama} and John {\sc Constable}
\vspace{0.5cm}
\\
{\normalsize\sl Faculty of Integrated Human Studies,}\\
{\normalsize\sl Kyoto University, Kyoto 606-8501, Japan}\\
{\normalsize\tt aoyama@phys.h.kyoto-u.ac.jp}\\
{\normalsize\tt john@ic.h.kyoto-u.ac.jp}\\
}

\date{\normalsize January, 1998}

\maketitle
\thispagestyle{empty}
\vfil
\vfil
\end{titlepage}

\renewcommand{\baselinestretch}{1.2}
\baselineskip=\baselinestretch\baselineskip

\def\topfraction{0.8}
\def\textfraction{0.1}
\def\floatpagefraction{0.8}

\newenvironment{FRAME}{\begin{trivlist}\item[]
\hrule \hbox to \linewidth\bgroup
\advance\linewidth by -30pt \hsize=\linewidth
\vrule\hfill \vbox\bgroup \vskip15pt
\def\thempfootnote{\arabic{mpfootnote}}
\begin{minipage}{\linewidth}}{
\baselineskip=\baselinestretch\baselineskip
\end{minipage}\vskip15pt
\egroup\hfill\vrule \egroup\hrule\end{trivlist}}

\newpage
\vspace*{20mm}
\begin{abstract} 
\normalsize
Recent observations in the theory of verse and empirical metrics 
have suggested that constructing a verse line involves a 
pat\-tern-matching search through a source text, 
and that the number of found elements (complete words totaling a 
specified number of syllables) is given by dividing the total number of 
words by the mean number of syllables per word in the source text.
This paper makes this latter point explicit mathematically, and
in the course of this demonstration shows that 
the word length frequency totals in English output are distributed 
geometrically (previous researchers reported an adjusted Poisson 
distribution), and that the sequential distribution is random at the 
global level, with significant non-randomness in the fine structure. 
Data from a corpus of just under two million words, and a syllable-count 
lexicon of 71,000 word-forms is reported. 
The pattern-matching theory is shown to be internally coherent, and it is 
observed that some of the analytic techniques described 
here form a satisfactory test for regular (isometric) lineation in a 
text. \\

\centerline{\sl Keywords: word length, metrics}
\end{abstract}


\newpage
\section{Introduction}
The making of metrical verse lines is a pattern-matching 
exercise, and in this respect contrasts sharply with the generation of
ordinary language output.
The composer must first generate source text, and then search through
it for elements which form, or may be accumulated to form, 
the desired pattern. This pattern is usually specified in relation 
to a selection from the available surface features of the language,
those in English verse, for example, being 
based principally on the patterning of stressed and unstressed 
syllables to form arrangements of beats and offbeats (\cite{Attridge1}, 
\cite{Attridge2}). 
In earlier work Constable (\cite{jc}) has observed that these 
rules lead
to the implication of a simpler and more readily studied rule, of which a
composer is usually unaware, namely that a line should consist 
of complete words totaling $n$ syllables, where $n$ can be a range. 
This may be put as a rule thus:

\begin{FRAME}
{\it `In every consecutive section of $n$
syllables there must be only complete words'}.
\end{FRAME}

The composition of word strings to fit a line definition rule of this 
type 
involves finding sequences of complete words totaling $n$ syllables,
or constructing them from sequences of less than $n$ syllables. 
In either case the activity is a pattern-matching 
search for a target, and the frequency of the target in the source 
language is of crucial interest to the composer, since a smaller 
number of found elements will more severely restrict the 
communicative options open. 
That is to say, the more targets are present in the 
source text, the more probable it is that the composer will find 
pieces that function adequately with regard to his or her 
purpose. Therefore, in order to facilitate line composition, 
authors are expected to take whatever action they can to increase 
the frequency of target elements. 
An effective way of doing this is to reduce the 
mean number of syllables per word in the source language being searched. 
Recent work in empirical metrics (\cite{jc}: 182) has demonstrated the 
relation between mean word 
length and target frequency empirically from small samples of 
English prose, and has claimed that the average frequency of a 
target object (a sequence of words totaling $n$ syllables) is 
given by the total number of words divided by the mean number of syllables
per word (and this leads to the prediction that when composing in verse 
authors
will tend to choose shorter words than they would otherwise have selected,
a point also explored in (\cite{jc}).
The mathematical reformulation required to substantiate and 
explain the factors underlying this effect involves detailed observations 
regarding the frequency and sequential word length distributions
typical of English, and we will now turn to this task.

\section{Global Structure}
In approaching word-length data, it is of crucial importance to
form an analytic strategy suitable for the purpose in hand. 
Given a distribution, either of frequencies or probabilities,
it is possible to fit it with any degree of accuracy to a function, 
as long as one has a large number of functions, each with a large
enough number of parameters. 
Previous research on word length frequency 
undertaken by the G\"ottingen Word Length Project (see \cite{best} for an 
overview of the project and 
a bibliography; prominent papers in the project include: \cite{becker}, 
\cite{dittrich}, \cite{frischen}, \cite{riedemann}, \cite{rottger}, 
\cite{wkga}, \cite{wimmer}, \cite{ziegler}, \cite{zuse}) 
has collected data relating to word length frequencies, usually from 
rather small samples, and then used software, the Altmann Fitter, to 
compute a best fit description, which in most cases proves to be an 
adjusted Poisson distribution (\cite{best}).
Such a fitting procedure is guaranteed to work:
the number of words with more than seven syllables is small, and they are 
infrequent
in output, so there are, normally, only approximately seven data points, 
and,
therefore, it will not be difficult to find an accurate fit if we have 
several hundred functions, 
each with several parameters, from which to choose.
Mathematically, if there are seven data items, 
a general function with 7 parameters would suffice 
for a {\it perfect} fit. 

We stress that this is not relevant: Although such a fit may serve 
certain 
technical purposes, it does not bring any insight into or understanding 
of the nature of the language under consideration.  It is only when we 
can find a fit with a much smaller number of parameters than that of the 
data, that we can move towards abstraction and understanding. Any 
deviation from this simple ansatz should be regarded as a subtle 
variation 
from the basic finding. 

Methodology of this type is quite common in exact science.
For example, in high-energy physics, where the purpose of 
experiments, say proton-electron collisions, is not to fit the 
experimental data with one of several thousand functions, 
but to grasp the nature of protons and their constituent quarks.
Once we know the fundamental properties, further exploration of
any deviation in the experimental data opens the way to advance
our understanding.

Our research is directed with this principle in mind, 
and we have therefore chosen not to fit the distributions using a large 
set of 
functions and parameters. 
Rather, we will first extract a simple property that describes the 
overall, global, structure of the data, from which we aim to obtain a 
deeper knowledge of the English language, if only in its syllabic 
organization. With this in hand we can then proceed to the study 
of deviations from this global structure, that is to the fine structure 
of the data.

\subsection{The Corpus}
We have analyzed the word-length structure of almost two million words
of prose by various authors as listed in Table \ref{tbl:authors}.
The texts were chosen with no other view than convenience, 
Constable having marked them in the course of other research. 
We acknowledge that a more principled choice would be 
desirable, and indeed hope to undertake such work ourselves, 
but we believe 
that this corpus is sufficient for present purposes. 
The syllabic data were obtained by using a simple marking program, 
constructed by Constable, which reads text and uses a custom-built 
lexicon to determine the syllabic count of each word form. 
When new word forms are encountered, the program requests human 
intervention, and the form is added to the lexicon. Consistency of 
syllable counting was ensured by the fact that only one user 
(Constable) has been responsible for building the lexicon, which 
at the time of writing contains 71,666 items. 
Abbreviations were expanded, and numbers were counted as if they were 
pronounced as hyphenated words 
(1,920 = one-thousand-nine-hundred-and-twenty), with the exception 
of years and dates which were counted in their normal pronounced 
form (1920 = nineteen twenty). None of these categories are, as it 
happens, frequent in our corpus.

\subsection{Frequency and Probabilities}
In order to present the method of analysis in definite terms,
let us introduce several mathematical notations and facilities.
We denote the syllabic data obtained as described above 
by a series of integers $N_i$ ($i=1,2,3,\cdots I$),
where $N_i$ is the number of syllables in the $i$-th word
and $I$ the total number of words in the data.

The number of sequences in a series $N_i$ which satisfy a line definition 
rule
of the type introduced in the
previous section is represented as follows;
\begin{equation}
n = \sum_{i=\ell}^{m \; ({\rm mod} \; I)} N_i.
\label{eqn:ld}
\end{equation}
The upper limit of this sum
implies that the line definition rule is applied with a `periodic 
boundary condition'; namely, the data is treated as a circle by
connecting the end of data with the beginning. 
This is a technical definition justifiable by its utility in 
the following mathematical treatment. Alternatively, one could use
a Dirichlet boundary condition, in which one simply terminates the
data sequence at $i=I$.  These boundary conditions, however, do not
significantly affect the results as long as the data size is
large, which is true for all the data we have analyzed.

It is straightforward to count the
number $L_{n,k}$ which match Eq.(\ref{eqn:ld}) for $k = m-\ell$ words:
The numbers $L_{n,1}$ are obtained simply by counting the numbers
equal to $n$ among the series $(N_1, N_2, \cdots, N_I)$.
Next, the numbers $L_{n,2}$ are obtained by counting similarly for
$(N_1+N_2, N_2+N_3, \cdots, N_I+N_1)$, 
$L_{n,3}$ from $(N_1+N_2+N_3, N_2+N_3+N_4, \cdots, N_I+N_1+N_2)$, 
and so forth.
By definition, the following identity is satisfied:
\begin{equation}
\sum_{n=1}^\infty L_{n,k} = I.
\label{eqn:lnorm}
\end{equation}
Since there are no zero-syllable words in English,
\begin{equation}
L_{n,k}=0 \quad \mbox{if} \quad n<k.  
\end{equation}
The quantity we are interested in is the number of sequences 
matching the line 
definition rule for {\it any} number of words, which is given 
by the following:
\begin{equation}
L_n = \sum_{k=1}^{n} L_{n,k}\, .
\label{eqn:lndef}
\end{equation}
This counting algorithm has been coded in {\it Mathematica} by Aoyama
and has been found to work much faster than
the original algorithm used by Constable (\cite{jc}:181).
In Table \ref{tbl:match} we give the partial list of $L_{n,k}$
and $L_n$ obtained for all the data listed in Table \ref{tbl:authors}.

For theoretical reasons it is best to deal with
quantities independent of the data size.
Therefore, we introduce the following normalized quantities:
\begin{equation}
P_{n,k} \equiv {L_{n,k} \over I}, \quad Q_n \equiv {L_n \over I} \, .
\end{equation}
Due to the identity (\ref{eqn:lnorm}), the following is satisfied:
\begin{equation}
\sum_{n=1}^\infty P_{n,k} = 1.
\label{eqn:pnorm}
\end{equation}
In this sense, a set of $P_{n,k}$ of a given $k$ defines
a probability distribution.
On the other hand, $Q_n$ does not have this property.
We call $Q_n$ a (normalized) `frequency'.
Corresponding to Eq.(\ref{eqn:lndef}), we have the following relation:
\begin{equation}
Q_n = \sum_{k=1}^{n} P_{n,k}\, .
\label{eqn:qndef}
\end{equation}
In Fig.\ref{fig:flat} we give the plot of $P_{n,k}$ and $Q_n$ for
all the data listed in Table \ref{tbl:authors}.
As is seen in Fig.\ref{fig:flat}, the most remarkable global feature of
the frequency distribution $Q_n$ is its flatness, that is,
its independence from $n$.
In order to analyze this structure, we may define an idealized constant
distribution $\bar Q_n=q$, where $q=0.720316$ is the average value of 
the actual distribution $Q_n$ for $n=1 \sim 30$.
In the following section we will examine what lies behind this constant 
distribution.

\subsection{Random-Ordering Hypothesis}
As a working hypothesis we might assume that the word-length series is
randomly ordered, or more accurately, that
\begin{FRAME}
{\it `the number of syllables in a word is independent of the number of
syllables in preceding words.'}  
\end{FRAME}
\noindent
In other words, this hypothesis suggests that there is no correlation
between the syllable-count values in the data series.
This random-ordering hypothesis allows us 
to express $P_{n,k}$ in terms of $P_{n,1}$,
the probability of a word having $n$ syllables (hereafter we denote
this quantity by $p_n = P_{n,1}$).
For example, two consecutive words
that satisfy the two-syllable line definition rule
can be obtained by having two one-syllable words in a row.
The number of one-syllable words is $Ip_1$, and according to the
random-ordering hypothesis the probability
of having a one-syllable word after a one-syllable word is
not affected by the first word having one-syllable , and therefore is 
$p_1$.
Thus the number of 2-syllable lines of this form
is given by $Ip_1 \times p_1$.
Dividing this by the total number of words, $I$, we obtain the
normalized frequency $P_{2,2}$:
\begin{equation}
P_{2,2} =  p_1^2.
\end{equation} 
For larger $n$ and $k$, combinatoric considerations must be 
addressed. For example, a three-syllable line can be created 
by having a two-syllable word and a one-syllable word in sequence, 
or vice versa. Counting all possibilities, we obtain,
\begin{equation}
P_{3,2} =  2 p_1 p_2. 
\end{equation}
Some of the other relations are listed in Table \ref{tbl:random}.
The general expression for $P_{n,k}$ can be obtained in a straightforward
manner, but is complicated in written from, and 
can be handled most simply by the use of generating functions.

We define a generating function $P_k(x)$ to represent
$P_{n,k}$ for $n=1 \sim \infty$ by the following:
\begin{equation}
P_k(x) \equiv \sum_{n=1}^\infty P_{n,k} \, x^n.
\end{equation}
Knowledge of all $P_{n,k}$ is equivalent to knowing the behaviour
of $P_k(x)$ near the origin $x=0$, as $P_{n,k}$ can be expressed 
as the $n$-th order derivative of $P_k(x)$ at $x=0$:
\begin{equation}
P_{n,k} = {1 \over \, n! \,} {\, d^n P_k \over dx^n}(0).
\label{eqn:qneq}
\end{equation}
The normalization condition (\ref{eqn:pnorm}) of $P_{n,k}$ 
is expressed as $P_k(1)=1$.
The various moments of $n$ (expectation values of powers of $n$)
can be expressed in terms of derivatives of $P_k (x)$ at $x=1$.  
For example, the average $\langle n \rangle$ 
and the standard deviation $\sigma$ of $n$ are given by the following:
\begin{eqnarray}
\langle n \rangle_k &\equiv& \sum_{n=1}^\infty n P_{n,k} = P_k^\prime(1),
\label{eqn:genave}\\
\sigma_k &\equiv& \sqrt{\langle (n - \langle n \rangle_k)^2 \rangle_k}
= \sqrt{\langle n^2 \rangle_k - \langle n \rangle_k^2}\nonumber\\
&=& \sqrt{P_k^{\prime\prime}(1) + P_k^\prime(1) 
- (P_k^\prime(1))^2}.
\label{eqn:gendev}
\end{eqnarray}
We also define a generating function $Q(x)$ as follows:
\begin{equation}
Q(x) \equiv \sum_{n=1}^\infty Q_n x^n.
\label{eqn:qngen}
\end{equation}
In terms of these generating functions, the relation 
(\ref{eqn:qndef}) is written as 
\begin{equation}
Q(x) = \sum_{k=1}^\infty P_k (x).
\end{equation}
A relation similar to Eq.(\ref{eqn:qneq}) holds also for $Q_n$.

The general expression of $P_{n,k}$ in terms of $p_n$ induced by the
random ordering condition can be summarized very simply. 
In terms of the generating functions it is expressed as follows:
\begin{equation}
P_k(x) = P_1(x)^k.
\label{eqn:simple}
\end{equation}
We note that the normalization is trivial in the above equation:
$P_k(1)= P_1(1)^k = 1$.
The relation (\ref{eqn:simple}) leads to 
the following expression of the generating function $Q(x)$: 
\begin{equation}
Q(x) = \sum_{k=1}^\infty P_k(x) 
= \sum_{k=1}^\infty P_1(x)^k 
= {P_1(x) \over 1 - P_1(x)}.
\label{eqn:qnfx}
\end{equation}
Thus the reason for introducing the random-ordering hypothesis
becomes evident:
If that hypothesis is valid, the features of the frequency distribution 
$Q_n$
can be explained by the features of the one-word probability distribution 
$p_n$ by using the relation (\ref{eqn:qnfx}).

We will now turn to the verification of the random-ordering hypothesis.
In Table \ref{tbl:corr} we list the number of $m$-syllable words
following immediately after $n$-syllable words.
The corresponding probability distribution is plotted in 
Fig.~\ref{fig:true}.
From this figure, we readily observe that this distribution is
almost independent from the value of $m$.
Therefore we confirm that the random-ordering hypothesis is valid to
a reasonable degree of accuracy.$^{1)}$

\subsection{Single Word Probability}
Now that the random-ordering hypothesis is confirmed, we can obtain the 
probability
$\bar p_n$ that induces the constant frequency distribution $\bar Q_n = 
q$.
The generating function for $\bar Q_n$ is as follows:
\begin{equation}
\bar Q(x) = \sum_{n=1}^\infty q x^n = {q x \over 1-x}.
\end{equation}
By solving Eq.~(\ref{eqn:qnfx}) in terms of $P_1(x)$, we obtain,
\begin{equation}
\bar P_1(x)={\bar Q(x) \over 1 + \bar Q(x)} = {q x \over 1- (1-q)x}
= \sum_{n=1}^\infty q (1-q)^{n-1} x^n \, .
\end{equation}
Therefore,
\begin{equation}
\bar p_n= q (1-q)^{n-1},
\label{eqn:geo}
\end{equation}
which is the geometric probability distribution. 
In Fig.~\ref{fig:geo} we compare the actual
probability distribution $p_n$ (dots) and the 
geometric distribution $\bar p_n$ in Eq.~(\ref{eqn:geo}) 
(dash-dotted line). As is seen in this plot, 
the agreement is close.
The geometric distribution (\ref{eqn:geo}) yields the average 
number of the syllables per word (mean word length) as follows:
\begin{equation}
\langle n \rangle = \sum_{n=1}^\infty n \bar p_n = \bar P_1'(1)
= {1 \over \, q \, }\, .
\end{equation}
This relation was observed earlier by Constable (\cite{jc}:182).
We stress that our new finding of the relation between the constant 
distribution $\bar Q_n$ and the geometric distribution $\bar p_n$,
which we reached through the application of the random-ordering hypothesis
gives a sound theoretical basis to this observation.

\subsection{Interpretation: Random Segmentation}
The two global properties we have found above, 
the random-ordering and the geometric distribution (\ref{eqn:geo}), 
allow a definite characterization of the word-length data, since these
are the properties typical of a system with a given probability of 
termination at any point: 
namely, if one assumes that sequences of syllables are 
constructed such that after any syllable, the end of a word happens 
with probability $q$, the above geometric distribution
(\ref{eqn:geo}) is obtained. Putting this in a slightly different 
manner, if one has a large number of syllables and word boundaries 
(spaces) with $(1-q)$ to $q$ ratio and randomly places them in 
sequence, the same distribution is obtained. 
We call this {\it random segmentation}.

The fact that this geometric distribution is not
found in the lexicon itself, which is plotted in Fig.~\ref{fig:dic},
has been noted by other researchers 
(\cite{wkga}), and provoked explanation in 
terms of attractors and control cycles in the composition process. 
Strictly speaking it is beyond the scope of our paper 
to engage deeply with this question. However, since we believe that these 
researchers have 
been misled by the presumption of order in the output distribution, there 
is some point in observing
that hypotheses based on randomness could, in principle, account for the 
relations
between these very different distributions, and explain the stability
of the geometric distribution in output.

For example, we might hypothesize that the concept of `word' or `word 
boundary' is a relatively late (though perhaps prehistoric) analytic
category, and has been arrived at by segmenting the verbal output
stream in such a way that word boundaries are placed with a fixed 
probability 
in
relation to syllable boundaries. The resulting word-forms are used to
compile a lexicon. Since the sound system of a language is not infinitely 
extendible, 
there will be more unique and acceptable disyllabic forms than 
monosyllables, more trisyllables 
than monosyllables, and so on. Thus, although in its early stages the 
lexicon would, obviously, 
follow the geometric distribution of the output it was drawn from 
(Fig.\ref{fig:geo}), 
eventually it would, temporarily, 
adopt the sort of curve seen in Fig.\ref{fig:dic}. 

This is not to suggest that the segmentation of English is fundamentally 
random, or that
`words' have a low linguistic reality, interesting though both 
speculations are. In line with the
data examined so far, our hypothesis merely notes that whatever principle 
of regular order
may be operating elsewhere, perhaps in relation to stress or phonemes, 
word boundaries and
syllable boundaries are related with a fixed probability.

\section{Fine Structures}
Readers will have noticed the differences between the global
structure and the actual distributions.
The most notable is the small dip of $Q_2$ below the
average value $q$ seen in Fig.\ref{fig:flat}, and 
the small differences between $p_{n,m}$ of different $m$ in
Fig.\ref{fig:true}.

When we discuss these differences,  
we need first to guard against statistical errors.
In other words, we first need to see whether these 
differences are meaningful quantities or can be attributed to 
statistical fluctuations.  Only when the former is more
likely, do we need to study the fine structures that explain these
differences.
We stress that this discussion of statistical errors is 
of the first importance.  
As we see below small data sets, such as those employed
by the G\"ottingen group, do in fact suffer from large statistical errors.
Detailed study of such data is either irrelevant or misleading.
In the following, we first discuss the handling of statistical
errors and then proceed to the discussion of features of the fine 
structure.

\subsection{Statistical Errors}
The standard estimate for statistical errors 
may be applied to the individual probabilities that we deal with 
this paper.
The 3-$\sigma$ error range for
the probability $P_{n,k}$ would be,
\begin{equation}
P_{n,k} - 3 \sqrt{P_{n,k}(1-P_{n,k}) \over I}
\sim 
P_{n,k} + 3 \sqrt{P_{n,k}(1-P_{n,k}) \over I}
\, .
\end{equation}
In other words, the true value lies in this range with
about 99.7\% probability.

The estimate of the error range for the frequencies
require more extensive discussion.
Such an estimate is made possible by relation (\ref{eqn:qndef}):
It is trivial for $Q_1$, as it is actually a probability, $Q_1 = P_{1,1}$.
Therefore its standard deviation is given by 
$\sigma = \sqrt{Q_1 (1-Q_1)}/\sqrt{I}$.
The next is $Q_2 = P_{2,1} + P_{2,2}$.
The probability $P_{2,1}$ is given by dividing the observed number 
$L_{2,1}$ of 2-syllable words by the total number of words $I$.
From Fig.\ref{fig:geo} we see that $P_{2,1} \sim 0.2$,
therefore the standard deviation of $L_{2,1}$ can be approximated as
$\sigma_{2,1} \sim \sqrt{L_{2,1}}$.
Similarly, we approximate that $\sigma_{2,2} \sim \sqrt{L_{2,2}}$.
Thus the total standard deviation for the observed number of strings 
$L_2$ is given as follows:
\begin{eqnarray}
\sigma_2^2 &=& \langle L_2^2 \rangle - \langle L_2 \rangle^2\nonumber\\
&=& L_{2,1} + L_{2,2} = L_2,
\end{eqnarray}
where we used the statistical independence of $L_{2,1}$ and $L_{2,2}$. 
That is, the standard deviation of $Q_2$ is 
given simply by $\sqrt{Q_2/I}$, just as if it is a probability by itself. 
The same is true for the rest of $Q_n$s.  

The estimate of the statistical errors for the
average value $q$ of $Q_n$ is simplified because of the flatness
of the distribution $Q_n$. Given the fact that the value of $Q_n$ is 
independent from $n$, the statistical errors follow from statistical
errors of any {\it one} value of $Q_n$, which is almost independent from 
$n$, as
explained above.
Therefore, we simply estimate the standard deviation of $q$ to be
of the same order as a typical value of that of $Q_n$s;$^{2)}$
Namely, we estimate the $3\sigma$-range of $q$ to be between $q \pm 3 
\sqrt{q/I}$.

\subsection{Deviations from the Flat $Q_n$ Distribution}
In order to study the small deviation of the frequency $Q_n$ 
from the flat distribution, we can plot the difference between the actual 
frequency
and the flat distribution itself, $\delta Q_n = Q_n - q$,
with vertical bars showing the 3 $\sigma$ 
error ranges, as in Fig.\ref{fig:sig}.

From this figure, we find that the 2 syllable depression 
plot is statistically significant, as are other deviations at
$n=1,3, 4$.
These fine deviations can be explained from underlying
deviations; namely, (1) deviation from the geometric distribution,
(2) deviation from random-ordering.
In order to examine these Fig.\ref{fig:nongeom} plots (a) $p_n - \bar 
p_n$ (solid line)
and (b) $p_{n,1} - \bar p_n$ (dotted line).
The line (a) is a measure of the deviation from the geometric
distribution, while the difference between (a) and (b) is a measure
of the deviation from random-ordering.
In these figures we find that
(1) monosyllabic words have a slightly higher probability
than that predicted by the geometric distribution,
while disyllabic words have a slightly lower probability,
and that (2) in the sequential distribution there is a slightly enhanced 
probability of a
polysyllable after a monosyllable (relative to random-ordering).
These are the important, indeed the only, exceptions to the
overall randomness in the distributions.
These deviations, being small compared to unity,
can be mathematically treated as perturbations in a 
randomly-ordered geometric distribution.
In that manner, it is straightforward to show that
these small deviations for smaller $n$ do not affect
$Q_n$ for large $n$, thus explaining the fact that deviation
of $Q_n$ from the flat distributin is localized to small $n$.

We have not examined the underlying linguistic causes of these 
deviations, and are not in
a position to do more than speculate. 
The corpus
is predominantly of high status literary writing, mostly 
nineteenth-century, and of that a substantial
portion comes from one author, Henry James. It might therefore be 
suggested that these
deviations are characteristic of an output type, or a period, or even of 
an author. However, some of the deviations observed overall are 
consistently found across authors and works, though
in the case of Kipling we found a significant depression at $n =1$ 
instead of an enhancement.
We predict, therefore, that some of these deviations, the depression
at $n = 2$ for example,
are universal characteristics, while some others will prove to be 
particular to an author, a work, a genre type, or a period. With regard
to the deviation from random sequencing, we suggest that the relation 
between commonly
occurring function terms, which are predominantly monosyllabic, and 
content terms, which
are somewhat more likely to be polysyllabic, is the likeliest
explanation.

Whatever the best causal account, it should be recognized that deviations 
such as these are
subtle variations from a strong fundamental trend, and it is not safe to 
conclude that they are
evidence which `confirm[s] the assumption of a non-accidental
distribution of word lengths' (\cite{ziegler}:73).

\subsection{Individual Authors}
The global structures we noted in the preceding section,
the flatness of the frequency $Q_n$ and the randomness of sequencing
are also true for individual authors and works.
In other words, it is not a result of averaging over large
fluctuations among authors.
In Fig.\ref{fig:flat-ge} and Fig.\ref{fig:true-ge}
we give the plot of the frequency $Q_n$ and the probability $p_{n,m}$
for George Eliot's novel {\it Middlemarch}. 
The global features are readily apparent from these figures.
The average frequency is $q=0.69844$ for the Eliot data and
is $q=0.720316$ for all the data in Table \ref{tbl:authors}, and readers 
may wonder whether this difference is meaningful.
As explained in a previous subsection, 
the standard deviation $\sigma_q$ of the average frequency $q$ is 
estimated to be $\sqrt{q}/I$.
This means that $\sigma_q \simeq \sqrt{0.720316/1977676} \simeq 0.00060$
for the corpus listed in Table \ref{tbl:authors}.
Thus at the $3\sigma$-confidence level, the true value of $q$
lies between $0.7185$ and $0.7221$. 
Similarly, the $3\sigma$-confidence range of $q$ for the George
Eliot data is $0.6940 \sim 0.7029$.
Since these ranges do not overlap, we conclude that indeed the mean 
number of syllables per word in Eliot is significantly 
larger than that of the corpus.
In Fig.\ref{fig:authorq} we give a plot of the values of $q$ and 
the $3\sigma$ ranges of all the authors and the whole corpus.

\subsection{Test for Lineation}
Apart from the $n=2$ deviation observable in the prose texts in our 
corpus, we are
aware of one large class of texts which routinely exhibit significant
deviations in the $Q_n$ distribution, namely isometrically lineated verse 
texts. 
This is hardly surprising. Texts composed in regular lines are by
definition ordered with respect to lineation, and this order 
will be detected by such a procedure as ours. If a poem is composed in 
lines of ten syllables, for example, then $n=10$, and all multiples
of ten, will be substantially above the flat distribution, and if
it is composed in two core line lengths, as limericks 
and Spenserian stanzas are, then it will exhibit two series of peaks.
Since we intend to discuss this matter at greater length elsewhere we 
will present only 
one example, the final version of William Wordsworth's {\it Prelude}, in 
Fig.\ref{fig:ww} and Fig.\ref{fig:wwd}. This poem, 
which was completed in 1839 but not published until 1850, is 
composed in blank verse, that is unrhymed five-beat lines in duple rhythm,
with a range of between 9 and 12 
syllables per line (in duple rhythm the offbeat position is usually a 
single syllable, but is sometimes
filled with two syllables, or even left unfilled). The poem contains 
7,849 lines and 57,570 words, with a mean number of syllables per word of 
1.4. 

Approximately 77.5 \% of the text is composed in ten syllable lines, 
with 19.4 \% being of eleven syllables and 2.3 \% of 
twelve syllables, together with a scattering of other lengths. This 
degree of concentration into the 
core line length is by no means abnormal, and if anything it is
somewhat more distributed than other texts examined. The $Q_n$
distribution reveals, apart from the expected depression at $n=2$,
significant peaks at ten and all multiples of ten, 
though the subsequent peaks are of course of lesser size, since 
long, uninterrupted, runs of ten-syllable lines are rare.

It should be noted that this test is of theoretical rather than practical 
value. It is unlikely that we will often wish to test in order to 
detect lineation, since the fact is usually visually evident, or clear 
from other features such as rhythm. 
However, as a contribution to the theoretical definition 
of verse lines and texts, particularly as distinguished from prose, the 
procedure is of 
considerable interest. 
Although it has been long obvious that lineation is not merely `a visual 
or typographical fact' but a `fact of the language' 
(\cite{wimsatt}:591), to use one 
well-known formula, there has been, to our knowledge, no conclusive 
empirical demonstration of the presence of this fact, or any
explanation of its character. The deviation from the flat distribution 
performs both these functions.

\section{Conclusion and Comments}

Previous research on word length 
distribution (\cite{becker}, \cite{best}, \cite{dittrich},
\cite{frischen}, \cite{riedemann}, \cite{rottger}, \cite{wkga}, 
\linebreak 
\cite{wimmer}, \cite{ziegler}, \cite{zuse}) has
attempted to infer significance from the non-geometric curve found,
and held that it supports the belief that `language is 
[...] a self-regulating system, which is controlled by the needs of the 
language community' (\cite{zuse}), or is an organism of interrelated 
control cycles (\cite{wkga}). Furthermore, these researchers have 
incautiously borrowed terms from
chaos theory, and been, in our view, misled by them. For example, in 
\cite{wkga}:98 it was 
claimed on the basis of a handful of data, that `the sequence of words 
is clearly chaotic', and that the distribution of word length in a text 
could be explained by reference to `attractors'. However, in the current 
context, there is in fact no chaos, in its mathematical sense, and what 
we observe in our study is 
randomness: when the sample size is 
small, any distribution, height or weight 
in a human population, or, to mention something fundamentally random, 
quantum theoretical events, will exhibit large fluctuations, and as the 
data size grows, the distributions become smoother.
We do not rule out the discovery of the sort of order sought by the 
synergetic linguists,
but observe that our findings give little support to its existence in 
relation to word length. 
Thus, in approaching frequency data of this type we find
ourselves generally in sympathy with those such as Mandelbrot (\cite{mandelbrot})
and Li (\cite {li}), who are
of course studying different linguistic features, in their advocation of
interpretations grounded in randomness, and we are less drawn to positions
such as those proposed by Zipf (\cite{zipf}:48), where statistical
regularities are seen to arise from some deep principle of order.

In conclusion, however, we should like to to emphasize that 
the theory and data outlined here are of more than negative value, or 
purely self-sufficing interest. Our 
investigation was derived from empirical observations and hypotheses 
offered in an earlier paper (\cite{jc}) with regard 
to the construction of verse lines, and those remarks are confirmed by 
our results. The relation between the mean number of syllables per
word and the number of sequences of words totalling a given number of 
syllables (\cite{jc}:182) is dependent on
the geometric frequency of word length totals, and the random
distribution of the words in the text sequence, which we have 
shown here to be solid findings. Thus, the apparently arcane facts of 
word length distribution in English output can be seen to deepen our
understanding of one, and a historically very important, area of
language output, isometrical verse.

\newpage

{\Large \bf Footnotes}

\begin{enumerate}

\item
We note that these features, flatness of the frequency $Q_n$ and 
random-ordering are also true for individual authors and works,
and are not a result of averaging over them.  These issues will
be addressed in subsequent sections.

\item
One might think that having a number of data points for $Q_n$ would 
reduce the 
standard deviation for $q$ by a factor equal to the square root of the
number of $Q_n$.  However, since the flat distribution has $\bar Q_n =q$
for $n=1 \sim \infty$, this argument would yield a zero standard 
deviation, which is clearly wrong.
\end{enumerate}

\newpage
\def\modosu{\hspace*{-7mm}}


\def\mycap#1{\caption[#1]
      {\protect\parbox[t]{11cm}{\baselineskip=16pt #1}}}
\newpage

\begin{table}
\begin{center}
\def\mb#1{\makebox[60pt]{#1}}
\def\arraystretch{1.2}
\begin{tabular}{|l|p{200pt}|r|} \hline
\makebox[90pt]{\sl Author} & \makebox[200pt]{\sl Title(s)} 
&\makebox[50pt]{\sl Words} \\ \hline
Bunyan, John & {\it Pilgrim's Progress} & 52,504 \\ \hline
Eliot, George & {\it Middlemarch} & 317,827 \\ \hline
Frankau, Gilbert & \baselineskip=17pt
{\it Woman of the Horizon} 
(section; first 67 pages 10 chapters) & 24,597 \\ \hline
Goldsmith, Oliver & {\it Vicar of Wakefield} & 63,096 \\ \hline
James, Henry & \baselineskip=17pt
{\it The Altar of the Dead; The Ambassadors; 
The American; The Aspern Papers; Confidence; Daisy Miller; 
Death of the Lion; The Europeans; The Figure in the Carpet; 
The Golden Bowl; An International Episode; Portrait of a Lady; 
Roderick Hudson; Sacred Fount; Turn of the Screw; Watch and Ward; 
Washington Square} & 1,285,041 \\ \hline
Kipling, Rudyard & {\it Rewards and Fairies; The Jungle Book} & 115,602 
\\ \hline
Milton, John & \baselineskip=17pt
{\it History of Britain; Colasterion; Martin Bucer} 
& 119,009 \\ \hline
\multicolumn{2}{|c|}{\sl Total} &1,977,676 \\ \hline
\end{tabular}
\end{center}
\mycap{Content of the corpus; authors, titles of the sources and 
number of words from each author}
\label{tbl:authors}
\end{table}
\newpage

\begin{table}
\begin{center}
\def\arraystretch{1.0}
\def\mb#1{\makebox[45pt]{#1}}
{\small
\begin{tabular}{|c|r|r|r|r|r|r|} \hline
$n$ & \mb{$L_{n,1}$} & \mb{$L_{n,2}$} & \mb{$L_{n,3}$} & \mb{$L_{n,4}$} 
& \mb{$L_{n,5}$} & \mb{$L_n$} \\ \hline
1  & 1,433,426 &       0 &      0 &      0 & 0 & 1,433,426 \\ \hline
2  &  371,500 & 1,025,719 &      0 &      0 & 0 & 1,397,219 \\ \hline
3  &  122,179 &  558,679 & 733,202 &      0 & 0 & 1,414,060 \\ \hline
4  &   40,314 &  246,132 & 611,737 & 531,686 & 0 & 1,429,869 \\ \hline
5  &    9,048 &   99,647 & 348,154 & 583,554 & 387,684& 1,428,087 \\ 
\hline
6  &    1,082 &   33,891 & 169,801 & 411,842 &  524,656& 1,425,780 \\ 
\hline
7  &     119 &    9,790 &  73,374 & 238,242 &439,613 & 1,426,115 \\ \hline
8  &       6 &    2,983 &  27,502 & 121,820 &293,091 & 1,426,874 \\ \hline
9  &       2 &     660 &   9,832 &  54,843 & 171,197 & 1,426,456 \\ \hline
10 & 0& 146 & 2,902 & 22,952 & 88,979 & 1,426,660 \\ \hline
11 & 0&  26 &   878 &  8,329 & 42,419 & 1,426,218 \\ \hline
12 & 0&   3 &   219 &  3,012 & 18,275 & 1,426,323 \\ \hline
13 & 0&   0 &    60 &    983 &  7,464 & 1,426,066 \\ \hline
14 & 0&   0 &    13 &    299 &  2,856 & 1,425,963 \\ \hline
15 & 0& 0& 2& 91& 941 & 1,425,162 \\ \hline
16 & 0& 0& 0& 16& 333 & 1,425,536 \\ \hline
17 & 0& 0& 0&  4& 108 & 1,425,480 \\ \hline
18 & 0& 0& 0&  2&  44 & 1,424,226 \\ \hline
19 & 0& 0& 0&  0&   8 & 1,424,392 \\ \hline
20 & 0& 0& 0&  1&   6 & 1,425,044 \\ \hline
21 & 0& 0& 0&  0&   1 & 1,425,327 \\ \hline
22 & 0& 0& 0&  0&   0 & 1,425,568 \\ \hline
23 & 0& 0& 0& 0& 1& 1,425,068 \\ \hline
24 & 0& 0& 0& 0& 0& 1,424,803 \\ \hline
25 & 0& 0& 0& 0& 0& 1,424,248 \\ \hline
26 & 0& 0& 0& 0& 0& 1,424,738 \\ \hline
27 & 0& 0& 0& 0& 0& 1,424,738 \\ \hline
28 & 0& 0& 0& 0& 0& 1,424,730 \\ \hline
29 & 0& 0& 0& 0& 0& 1,424,089 \\ \hline
30 & 0& 0& 0& 0& 0& 1,424,313 \\ \hline
\end{tabular}}
\end{center}
\mycap{Number of strings $L_{n,k}$ and $L_n$ 
that satisfy the $n$-syllable line definition rule for $n=1 \sim 30$ and 
$k=1 \sim 5$. The values of $L_{n,k}$ for $k=6 \sim 30$ are omitted
due to space limitations. These figures cover all two-million words of 
data listed
in Table \protect{\ref{tbl:authors}}.}
\label{tbl:match}
\end{table}
\newpage
\begin{table}[p]
\vspace*{30mm}
\begin{center}
\def\mb#1{\makebox[50pt]{#1}}
\def\arraystretch{1.3}
\begin{tabular}{|c|c|c|c|c|c|} \hline
  & \mb{$n=$1} & \mb{2} & \mb{3} & \mb{4} & \mb{5} \\ \hline
\makebox[30pt]{$P_{n,1}$} & $p_1$ & $p_2$ & $p_3$ & $p_4$ & $p_5$ \\ 
\hline
$P_{n,2}$ & 0 & $p_1^2$ & $2 p_1 p_2$ & $2 p_1 p_3 + p_2^2$ 
& $2 (p_1 p_4 + p_2 p_3)$ \\ \hline
$P_{n,3}$ & 0 & 0 & $p_1^3$ & $3 p_1^2 p_2$ 
& $3(p_1^2 p_3 + p_1 p_2^2)$\\ \hline
$P_{n,4}$ & 0 & 0 & 0 & $p_1^4$ & $4p_1^3 p_2 + 6 p_1^2 p_2^2$ \\ \hline
$P_{n,5}$ & 0 & 0 & 0 & 0 & $p_1^5$  \\ \hline
\end{tabular}
\end{center}
\mycap{Some of the consequences of the Random-Ordering Hypothesis.
The probability $P_{n,k}$ is listed at the ($k$, $n$) position.}
\label{tbl:random}
\vspace*{30mm}
\end{table}
\newpage
\begin{table}[p]
\vspace*{20mm}
{\small
\begin{center}
\def\mb#1{\makebox[30pt]{#1}}
\def\mbs#1{\makebox[15pt]{#1}}
\def\arraystretch{1.2}
\begin{tabular}{|c|r|r|r|r|r|r|r|r|r|} \hline
$m$ & \makebox[50pt]{$n=1$}  
  & \mb{2} & \mb{3} & \mb{4} & \mb{5} & \mbs{6} & \mbs{7} &
  \mbs{8} & \mbs{9} \\ \hline
all & 1,433,426 & 371,500 & 122,179 & 40,314 & 9,048 & 1,082 & 119 & 6 
  & 2  \\ \hline
1 & 1,025,719 & 279,915 & 90,473 &  29,731 & 6,680 & 811 & 91 
  & 5 & 1 \\ \hline
2 & 278,764 & 63,357 & 20,960 & 6,733 & 1,485 & 183 & 17 & 0 & 1 \\ \hline
3 & 92,302 & 19,542 & 7,263 & 2,422 & 589 & 52 & 8 & 1 & 0\\ \hline
4 & 29,414 & 6,815 & 2,710 & 1,128 & 217 & 27 & 3 & 0 & 0 \\ \hline
5 & 6,400 & 1,617 & 692 & 265 & 67 & 7 & 0  & 0 & 0 \\ \hline
6 & 745 & 225 & 71 & 30 & 9 & 2 & 0 & 0 & 0  \\ \hline
7 & 75 & 28 & 10 & 5 & 1 & 0 & 0  & 0 & 0 \\ \hline
8 & 5 & 1 & 0 & 0 & 0 & 0 & 0  & 0 & 0 \\ \hline
9 & 2 & 0 & 0 & 0 & 0 & 0 & 0  & 0 & 0 \\ \hline
\end{tabular}
\end{center}
}
\mycap{List of the number of occurrences of $n$-syllable words after
$m$-syllable words}
\label{tbl:corr}
\vspace*{20mm}
\end{table}
\newpage
\begin{figure}
\centerline{\epsfxsize=155mm \epsfbox{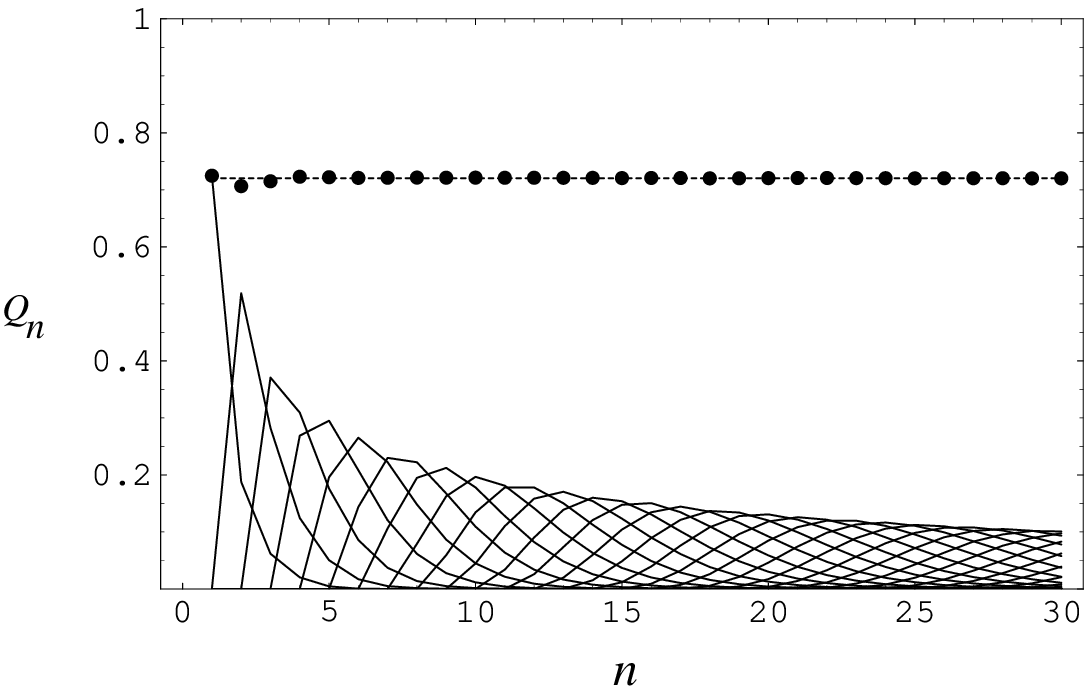}} \vspace*{-15mm}
\mycap{Plot of the normalized frequency $Q_n$ and the probability
distributions $P_{n,k}$ for the data in Table \protect{\ref{tbl:authors}}.
The horizontal dashed line shows the average value $q=0.720316$ for 
$Q_n$.}
\label{fig:flat}
\end{figure}
\newpage
\begin{figure}
\centerline{\epsfxsize=155mm \epsfbox{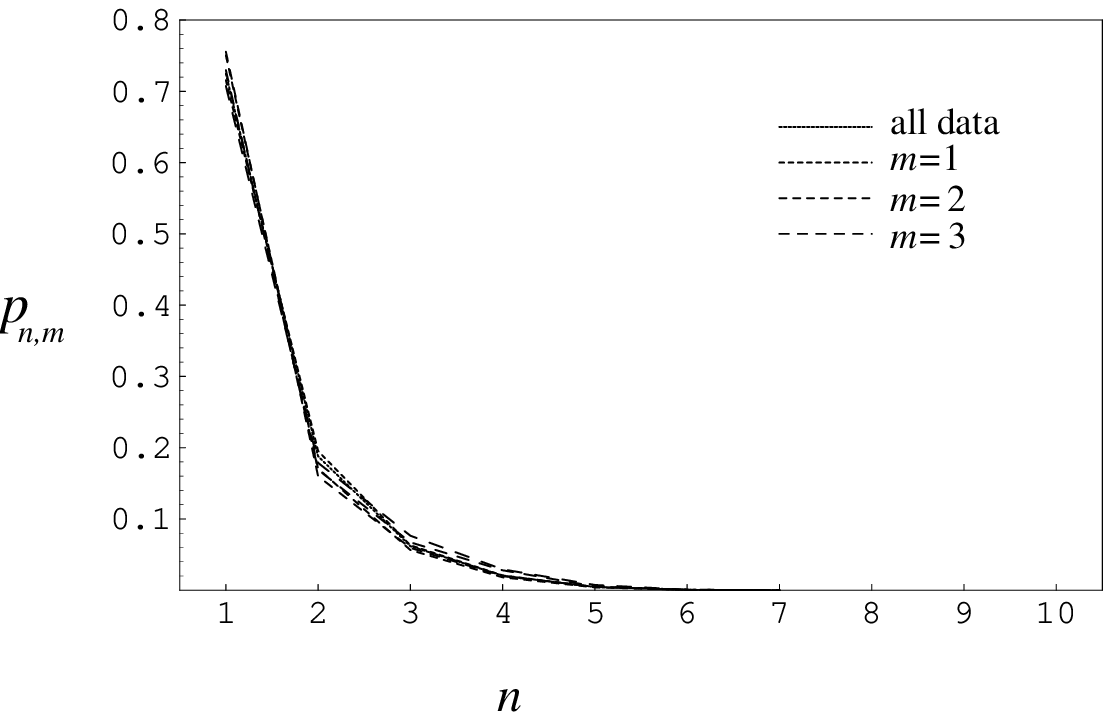}}
\vspace*{-10mm}
\mycap{Plot of the probability distribution $p_{n,m}$ for the data 
in Table \protect\ref{tbl:corr}. The solid line shows $p_n$, while
other lines show $p_{n,m}$.}
\label{fig:true}
\end{figure}
\newpage
\begin{figure}
\centerline{\epsfxsize=155mm \epsfbox{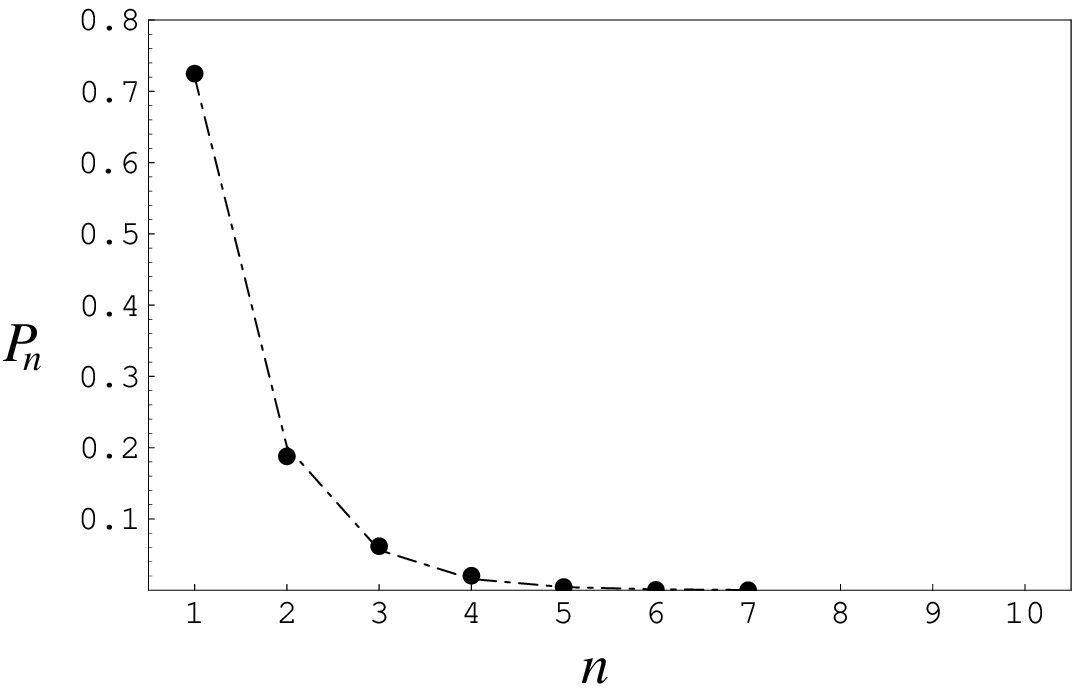}}
\vspace*{-10mm}
\mycap{The probability distribution $p_n$ (denoted by dots)
and the theoretical prediction $\bar p_n$ 
(\protect{\ref{eqn:geo}}) (denoted by the dash-dotted line)}
\label{fig:geo}
\end{figure}
\newpage
\begin{figure}
\centerline{\epsfxsize=155mm \epsfbox{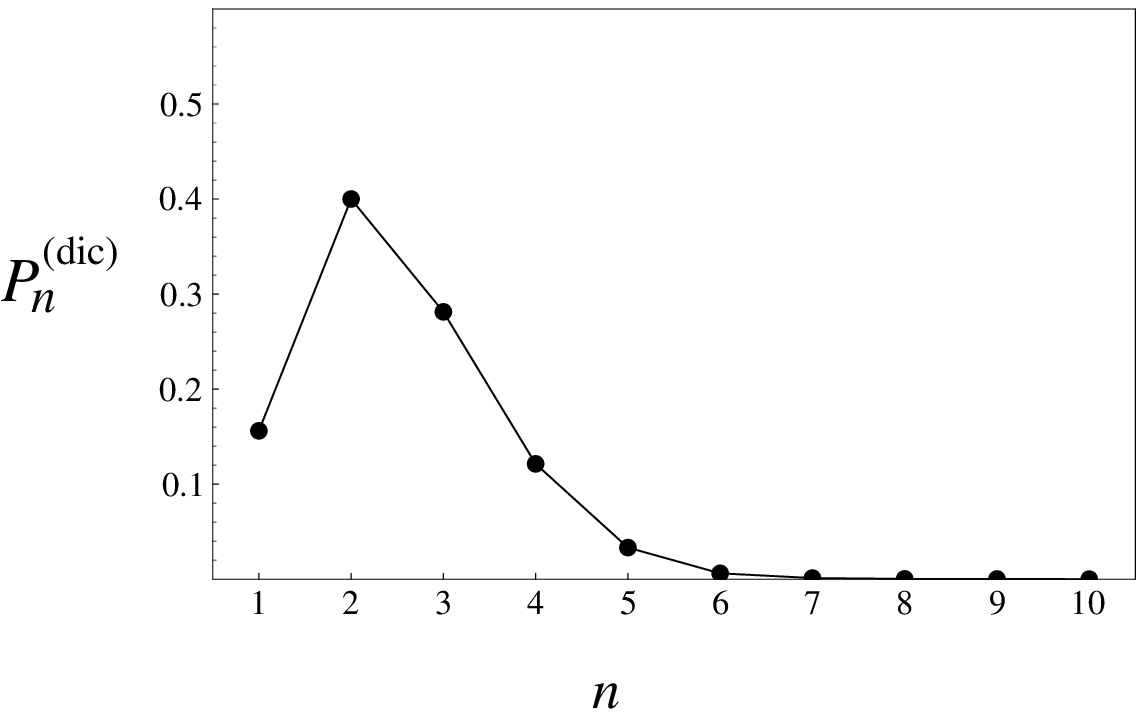}}
\vspace*{-10mm}
\mycap{Probability distribution for words with $n$-syllables 
in Constable's lexicon}
\label{fig:dic}
\end{figure}
\newpage
\begin{figure}
\centerline{\epsfxsize=155mm \epsfbox{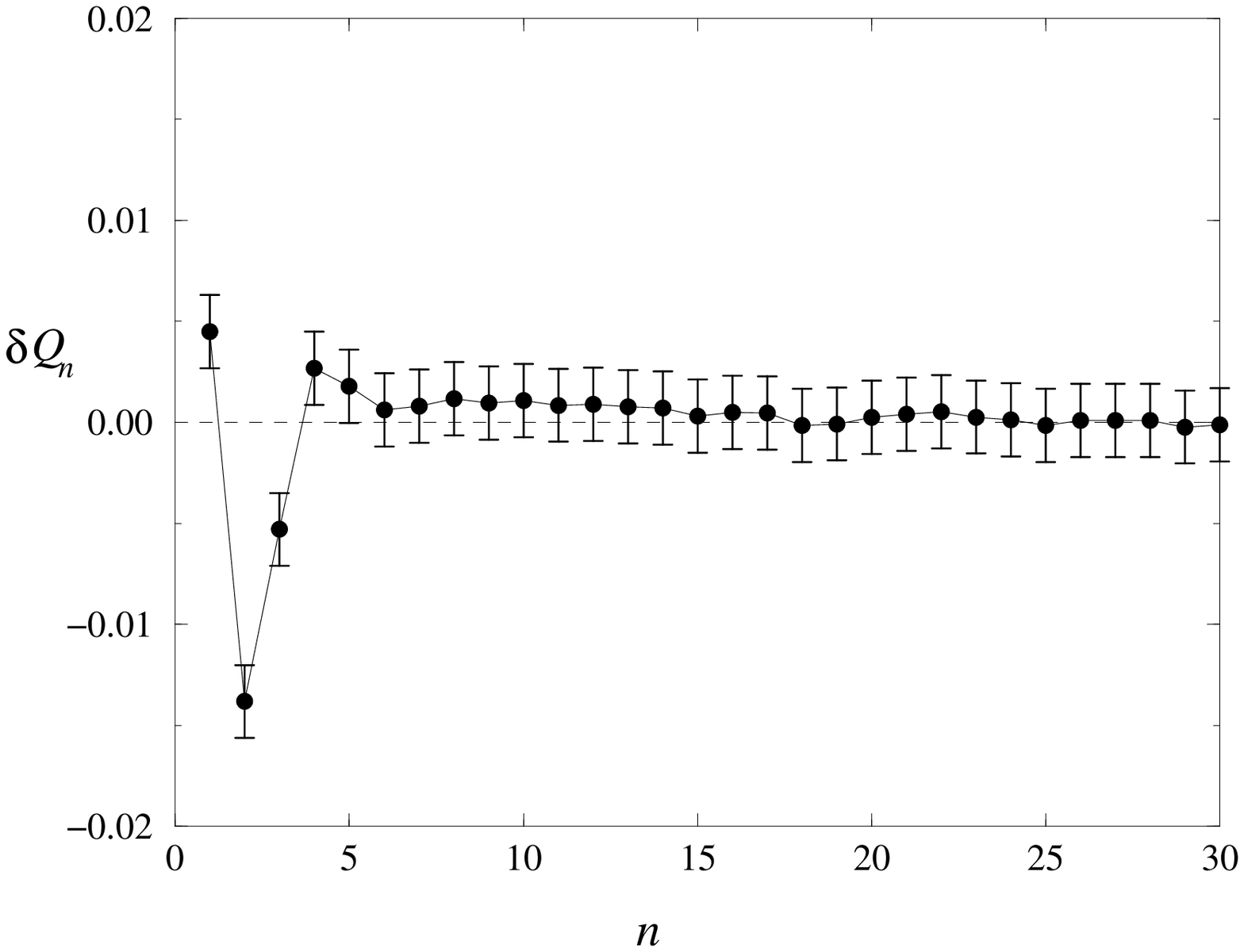}}
\mycap{Detail of $\delta Q_n = Q_n - q$.
The vertical bars show the 3$\sigma$-confidence ranges of 
statistical errors for each data point.
Note that the horizontal range covers only 1/25 of the range of  
Fig.\protect{\ref{fig:flat}.}}
\label{fig:sig}
\end{figure}
\newpage
\begin{figure}
\centerline{\epsfxsize=155mm \epsfbox{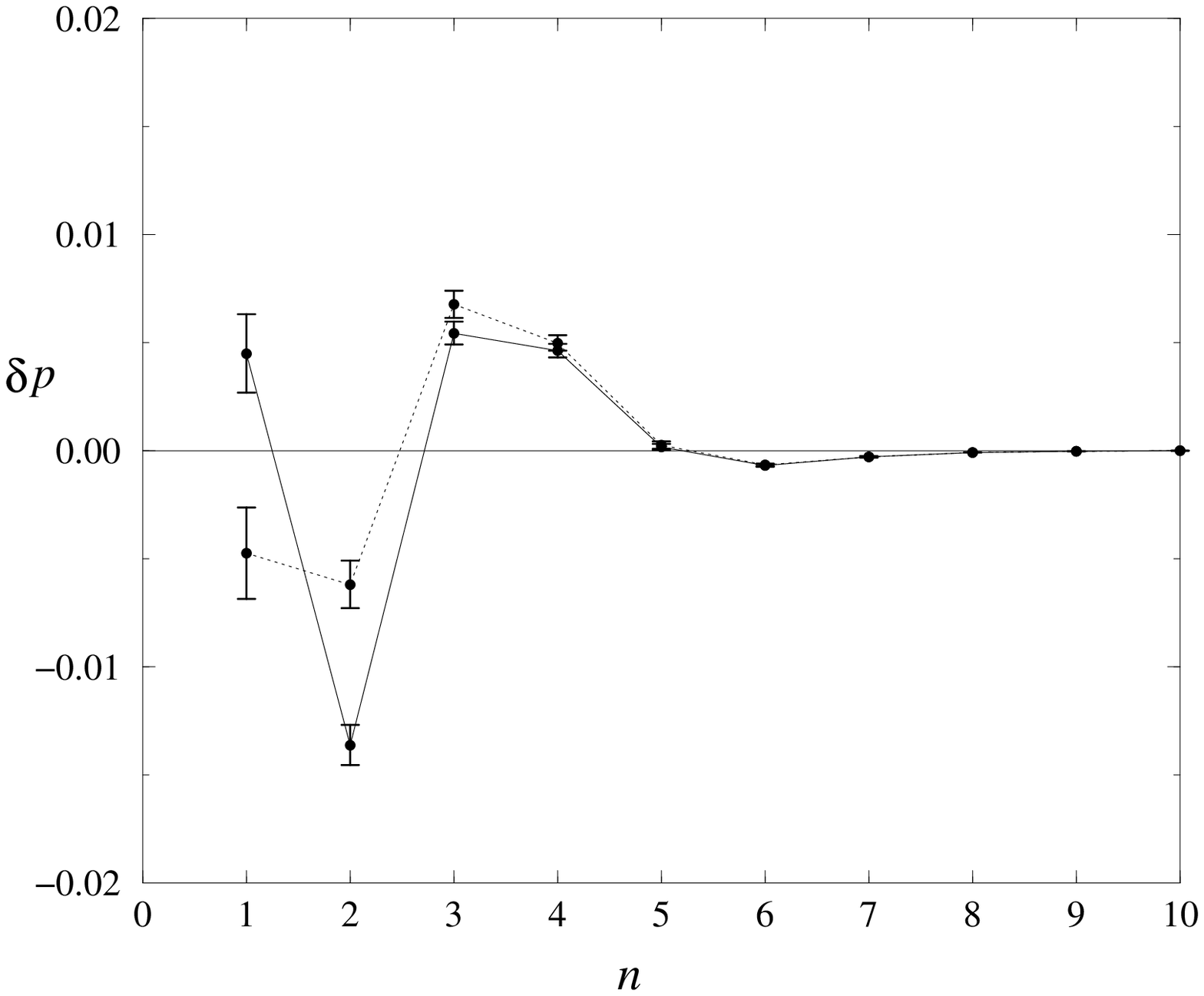}}
\mycap{Detail of (a) $p_n - \bar p_n$ (solid line)
and (b) $p_{n,1} - \bar p_n$ (dotted line)}
\label{fig:nongeom}
\end{figure}
\newpage
\begin{figure}
\centerline{\epsfxsize=155mm \epsfbox{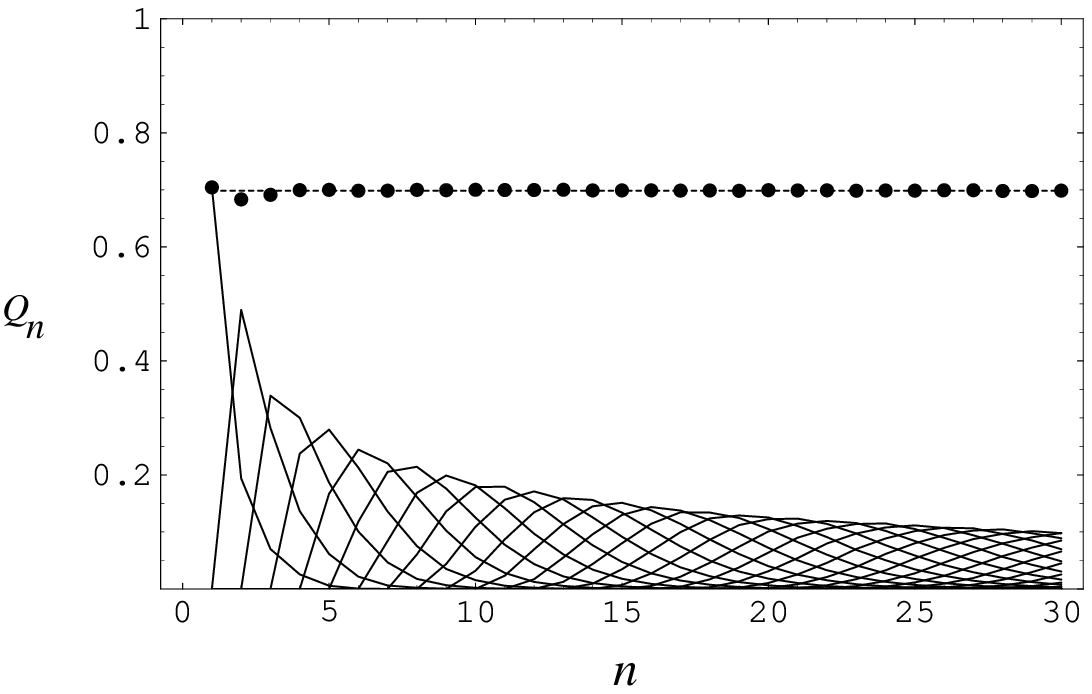}}\vspace*{-15mm}
\mycap{Plot of the normalized frequency $Q_n$ and the probability
distributions $P_{n,k}$ for George Eliot, {\it Middlemarch}.
The horizontal dashed line shows the average value $q= 0.69844$ for 
$Q_n$.}
\label{fig:flat-ge}
\end{figure}
\newpage
\begin{figure}
\centerline{\epsfxsize=155mm \epsfbox{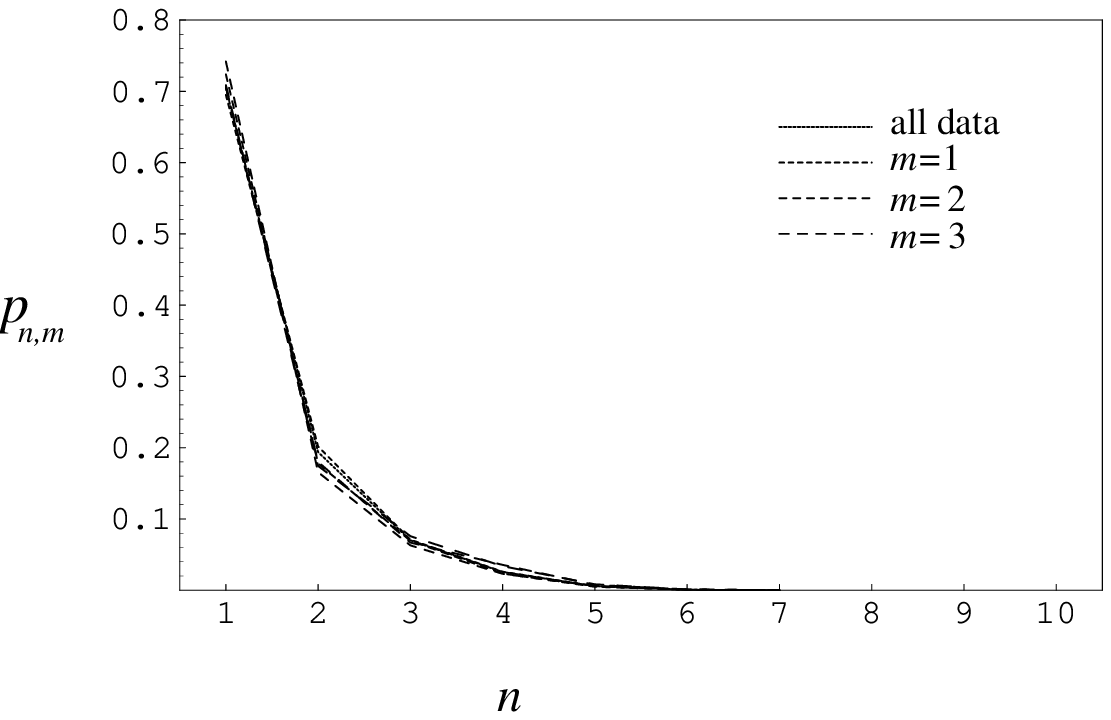}}\vspace*{-10mm}
\mycap{Plot of the probability distribution $p_{n,m}$ for George Eliot, 
{\it Middlemarch}}
\label{fig:true-ge}
\end{figure}
\newpage
\begin{figure}
\centerline{\epsfxsize=155mm \epsfbox{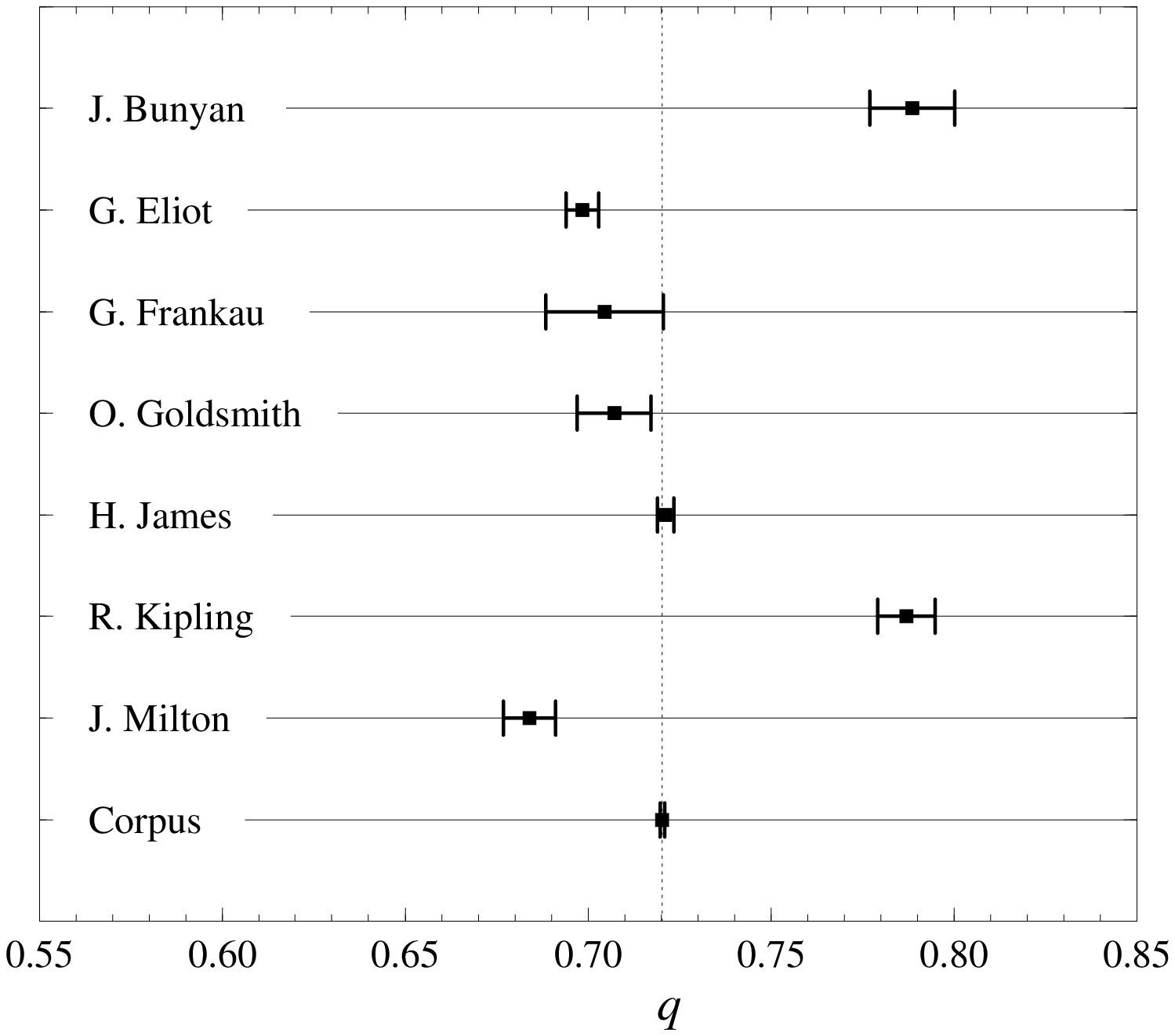}}
\mycap{Plot of the value of $q$ and 
its $3\sigma$ range for all the authors and the whole corpus}
\label{fig:authorq}
\end{figure}
\newpage
\begin{figure}
\centerline{\epsfxsize=155mm \epsfbox{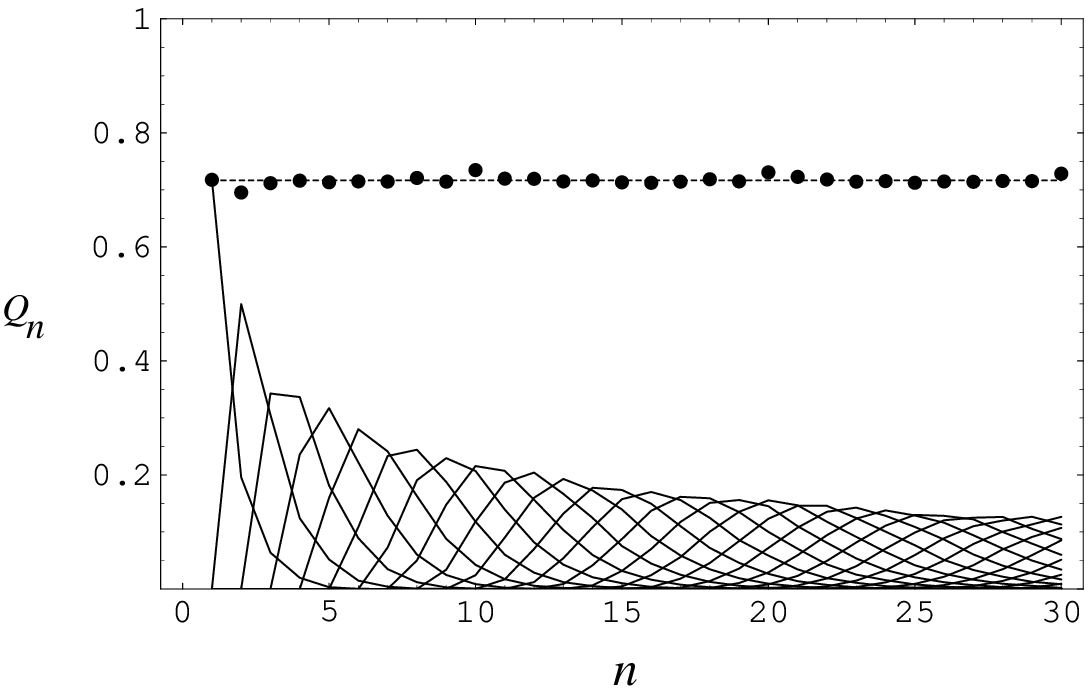}}\vspace*{-15mm}
\mycap{The frequency $Q_n$ and the probabilities $P_{n,k}$ 
for Wordsworth's Prelude (57,570 words)}
\label{fig:ww}
\end{figure}
\newpage
\begin{figure}
\centerline{\epsfxsize=155mm \epsfbox{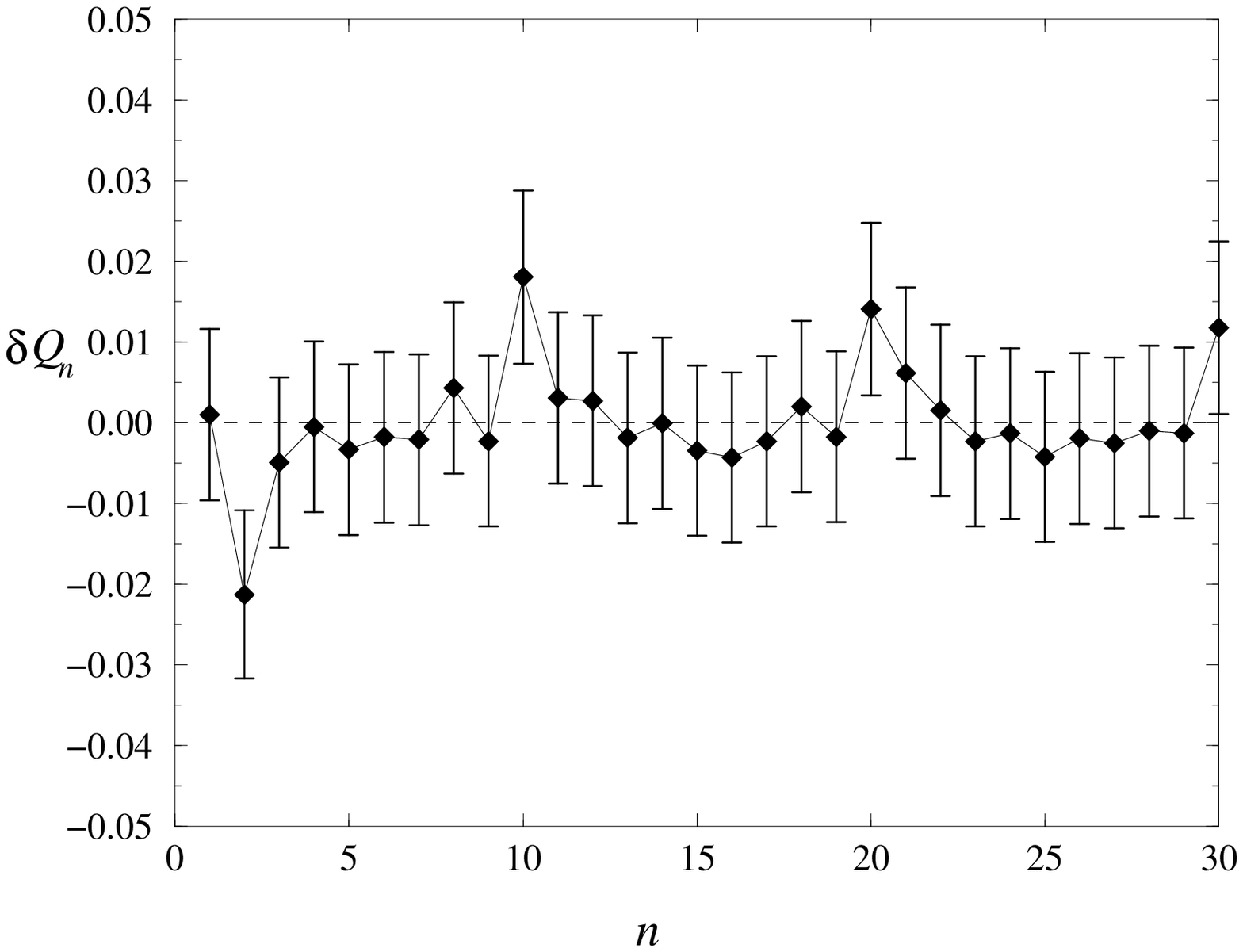}}
\mycap{Detailed view of the frequency $Q_n$ and the probabilities
  $P_{n,k}$ of Wordsworth's Prelude.}
\label{fig:wwd}
\end{figure}

\end{document}